\documentclass[letterpaper]{article} 
\usepackage[preprint]{aaai2027} 
\usepackage[hyphens]{url}  
\usepackage{graphicx} 
\usepackage{natbib}  
\usepackage{caption} 
\usepackage{amsmath}
\usepackage{amsfonts}
\usepackage{algorithm}
\usepackage{algorithmic}

\usepackage{newfloat}
\usepackage{listings}
\DeclareCaptionStyle{ruled}{labelfont=normalfont,labelsep=colon,strut=off} 
\floatstyle{ruled}
\newfloat{listing}{tb}{lst}{}
\floatname{listing}{Listing}

\usepackage{booktabs}

\title{When Self-Evolution Backfires: Pre-Commit Gating against Skill Contamination in LLM Agents}

\author{
    Linfang Shang\equalcontrib, Ming Xu\corresponding, Yiding Sun, Tianle Xia, Lingxiang Hu,\\
    Lan Xu, Ning Zheng
}
\affiliations{
    Tencent\\
    \{faelynshang, flemingxu, emanuelsun, tianlexia, lingxianghu, \\
    lanxu, yodazheng\}@tencent.com
}

\begin{document}

\maketitle

\begin{abstract}
Self-evolving agents accumulate capability by distilling reusable skills from their execution trajectories, but we find this process is not monotonic: past a critical pool size, newly added skills degrade performance instead of improving it. We formalize this capability--contamination phase transition and trace it to a structural cause---once a defective skill enters the decision context, it becomes reference material for distilling later skills, forming cross-round contamination chains. We further show the contamination is structurally irreversible: removing a source skill after the fact cannot erase the flawed reasoning its descendants have already inherited, so post-hoc rollback recovers only a small fraction of the lost performance. This makes skill admission a pre-commit necessity rather than a post-hoc fix, and motivates Verifier-as-Gatekeeper (VaG): a progressive trust hierarchy whose three heterogeneous critics---structural validity, behavioral harmlessness, and semantic consistency---filter each skill individually, coupled with a marginal-gain subset selection that removes combinatorial contamination at the top tier before skills reach the runtime context. On Terminal-Bench~2, unconditional accumulation rises to a peak and then degrades, giving back most of its gains as the pool keeps growing, and post-hoc removal of the culprit skills recovers only a small part of the drop---the empirical signature of irreversibility. In contrast, VaG improves every round, reaching 72\% pass@1 with a pool roughly $5\times$ smaller, and its frozen skill pool transfers positively to four other backbones and a second benchmark without re-evolution. Ablations confirm the three critics are complementary and mutually non-substitutable, each intercepting a largely disjoint class of harmful skills.
\end{abstract}


\section{Introduction}
A self-evolving agent improves by writing down what it learns. After each round
of task solving, its trajectories are distilled into short natural-language
skills, appended to a persistent pool, and injected into the decision context of
every later round
\cite{tao2024survey,
yao2022react,
park2023generative,wang2024survey,zhang2022automatic,schick2023toolformer}.
Capability is supposed to compound without touching model weights: more rounds,
more skills, a better agent.

Existing systems share one step in this loop that has escaped scrutiny:
\emph{whatever the distiller emits is admitted}. Beyond deduplication, no skill
has to justify its place in the pool before it starts steering behavior.
Admission looks like bookkeeping, and bookkeeping looks reversible---a bad entry
can always be deleted later. We find that neither part of this intuition holds.

\begin{figure}[t]
\centering
\includegraphics[width=\columnwidth]{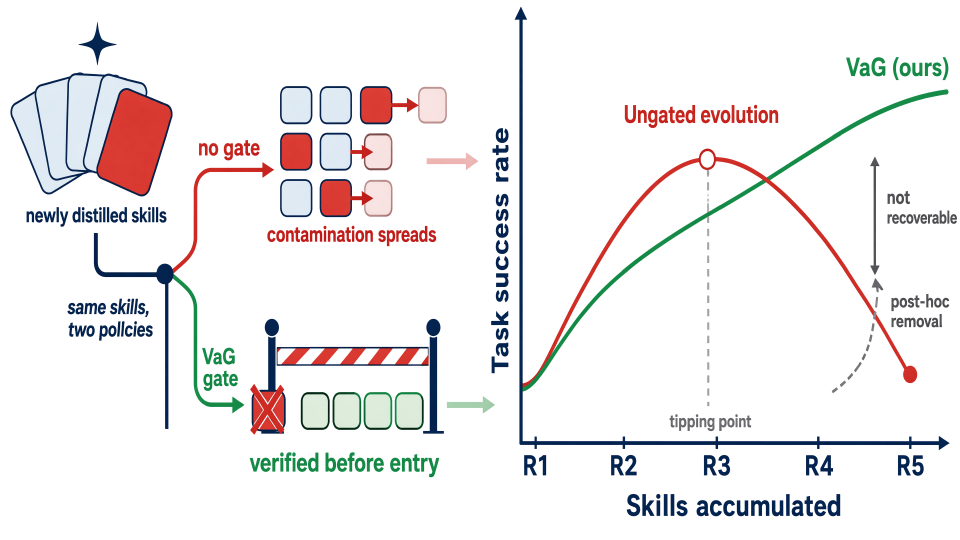}
\caption{
Unconditional skill admission makes evolution non-monotone.
As skills accumulate, an ungated agent improves, peaks at a critical pool size,
and then degrades, giving back most of its gains, because defective skills both
mislead the agent directly and become reference context for later distillation.
Removing the identified culprits afterwards recovers only part of the loss,
since their descendants remain.
VaG verifies each skill before it can enter the runtime context. It gives up a
little while the pool is small, since gating also rejects some benign skills, but it improves monotonically and ends above what ungated evolution reaches at its best---with a smaller, cleaner pool.
}
\label{fig:teaser}
\end{figure}

\paragraph{Skill accumulation is not monotone.}
Tracking an agent over five evolution rounds on Terminal-Bench~2~\cite{merrill2026terminal}, pass@1 rises
while the pool is small, peaks at a critical size $k^*$, and then falls back to
roughly its first-round level---even though every added skill was distilled from
a trajectory the agent itself had completed (Figure~\ref{fig:teaser}). We call
this peak the \emph{capability--contamination tipping point} and trace it to
three mechanisms: \emph{individual} contamination, where a single skill lowers
the success rate on its own; \emph{combinatorial} contamination, where skills
that are each harmless conflict once injected together; and \emph{systemic}
contamination, the macroscopic non-monotonicity the first two produce.

\paragraph{Deleting a bad skill afterwards does not undo it.}
Skills are distilled \emph{conditioned on} the pool that was live at the time,
so a defective skill admitted in round $r$ belongs to the reference context of
everything distilled after $r$. Its flawed reasoning survives in descendants
that never name it. Removing the source alone is therefore strictly weaker than
removing the source together with its entire lineage---and lineage cleanup is
out of reach in practice, because existing skill libraries do not record which
skills were in context when a given skill was written. Post-hoc curation carries
a recovery gap that better detection alone cannot close.

Taken together, these two observations change what skill admission \emph{is}. It
is not an optimization to reach for once the pool gets messy; it is the last
point in the loop at which contamination can still be stopped. Verification must
happen \textbf{before} a skill enters the runtime context.

\paragraph{Verifier-as-Gatekeeper.}
We therefore treat admission as an explicit promotion decision rather than a
write. Every distilled skill starts \emph{Cold}: recorded, but invisible to the
agent. Reaching the runtime context requires clearing two gates of increasing
cost. The first asks whether a skill is harmless \emph{on its own}, and answers
it with three deliberately heterogeneous critics: a deterministic schema check,
a single-skill A-B replay on held-out tasks, and one LLM review for fabricated
or self-contradictory advice. Because the three inspect disjoint properties
(format, behavior, meaning), a harmful skill has to fool all of them to slip
through, and the two cheap checks reject most candidates before any LLM is
called. The second gate asks whether a skill is harmless \emph{in company}:
promotion to the runtime tier is a greedy subset selection against measured
joint performance, which is the only way to catch skills that are individually
fine and jointly harmful. Because skill utility is not globally monotone---a
skill can \emph{lower} joint performance---we use a marginal-gain greedy that
adds a candidate only when it improves measured joint performance, directly
targeting the combinatorial conflicts that per-skill checks cannot detect.

\paragraph{Findings.}
On Terminal-Bench~2, ungated evolution reproduces the predicted tipping point,
peaking at round~3 and then losing most of that gain, ending at R5 only 2 points above its own R1,
whereas VaG improves across all five rounds and ends 10 points above ungated
evolution's \emph{best} round with a pool roughly 5$\times$ smaller---so gating
beats even oracle early stopping, not just unchecked accumulation.
Component ablations show that the three verification dimensions are
complementary rather than redundant: removing any one---and especially the
behavioral replay check---degrades performance and inflates the pool, because
each critic intercepts a largely disjoint class of harmful skills.
Source-only rollback recovers only 17\% of the degradation,
matching the irreversibility argument. A frozen VaG pool finally transfers to
other backbones without re-evolution, indicating that gated evolution distills
model-agnostic engineering experience rather than backbone-specific artifacts.

\paragraph{Contributions.}
\begin{itemize}
\item \textbf{Phenomenon.} We identify and formalize the
capability--contamination tipping point in self-evolving agents, with a
three-level taxonomy spanning individual, combinatorial, and systemic
effects.
\item \textbf{Irreversibility.} We show that contamination propagates through
derived-skill lineages and is structurally irreversible: source-only rollback
is dominated by full lineage cleanup, establishing a recovery gap that applies
to every post-hoc remediation strategy.
\item \textbf{Method.} From this necessity we derive VaG, a pre-commit gating
mechanism combining a progressive trust hierarchy of three heterogeneous
critics with a marginal-gain greedy selection that removes combinatorial
contamination at the top tier.
\item \textbf{Evidence.} On Terminal-Bench~2 we demonstrate the tipping
point, measure the rollback recovery gap, show monotone improvement under
gating, and verify cross-backbone transfer of the evolved pool.
\end{itemize}

\section{Related Work}

\paragraph{Self-Evolving Agents and Skill Distillation.}
Recent work lets agents accumulate capability by distilling reusable skills from execution traces for later reuse~\cite{wang2023voyager, lin2026agentic, zhang2025agentic, mei2026searcharttraininglonghorizonsearch,evolver,sage}.
Experiential agents such as ExpeL~\cite{zhao2024expel} and Agent Workflow Memory~\cite{wang2024agent} follow the same recipe, distilling past experience into natural-language insights or reusable workflows appended to the agent's context~\cite{packer2023memgpt,huang2023large,zelikman2022star,gulcehre2023reinforced}.
One design choice is shared but rarely questioned: a distilled skill is written into persistent storage unconditionally and takes effect at the next step.
Voyager~\cite{wang2023voyager} verifies only that a skill completes its target task before storing it, but never checks whether that skill degrades performance on \emph{other} tasks---precisely the failure mode we study.
MetaClaw~\cite{xia2026metaclaw} uses dual-timescale meta-learning with reinforcement fine-tuning, which requires access to model weights (infeasible for closed-source APIs) and still applies no gate at admission time.
AHE~\cite{lin2026agentic} evolves the full harness at scale, but its evaluation is deferred and operates at the harness level: each round scores only the aggregate harness and cannot attribute a change to any individual skill.
ACE~\cite{zhang2025agentic} scales context-level evolution further, yet inherits the same unconditional-admission assumption.
Our work departs on two axes: verification granularity moves from the harness to the individual skill, and verification timing moves from deferred post-evaluation to a pre-commit gate applied before a skill can enter the agent's runtime context.

\paragraph{Submodular Selection and Verification in Agent Pipelines.}
Submodular optimization offers classical approximation guarantees for combinatorial subset selection~\cite{nemhauser1978analysis,feige2011maximizing}, later extended to weak submodularity~\cite{das2011submodular} and to noisy oracles~\cite{mirzasoleiman2014lazier}.
Within agent systems, existing verification tends to be coarse-grained: self-refinement loops~\cite{madaan2023self, shinn2023reflexion, skillops, routeguard} critique a whole trajectory after execution, while Voyager-style libraries~\cite{wang2023voyager} apply at most heuristic deduplication when storing skills.
None of these address the combinatorial interaction we highlight: skills that are each individually harmless can still degrade performance when injected together through conflicting advice.
We bring marginal-gain subset selection into skill admission, using held-out estimates to guide greedy construction of the admitted set---a combinatorial safety check that prior agent-verification pipelines, which judge skills only in isolation, do not offer.

\begin{figure*}[t]
\centering
\includegraphics[width=\textwidth]{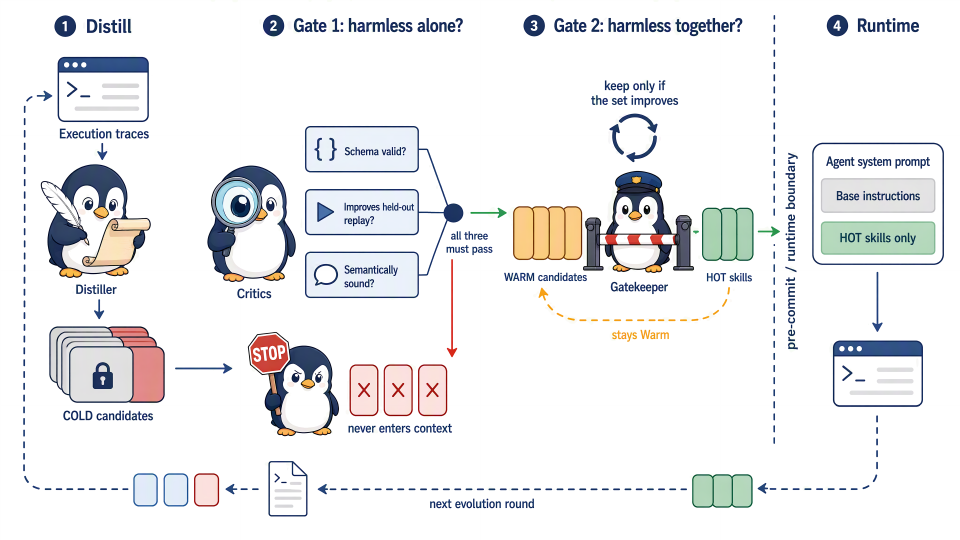}
\caption{
\textbf{The Verifier-as-Gatekeeper pipeline.}
(1)~Skills distilled from execution traces enter the Cold tier, invisible to the agent.
(2)~\emph{Gate~1} asks whether a skill is harmless \emph{on its own}: it must pass all three heterogeneous critics---schema validity, single-skill A-B replay on held-out tasks, and one LLM semantic review---or it never enters the context.
(3)~\emph{Gate~2} asks whether the survivors are harmless \emph{together}: greedy selection on measured joint held-out performance promotes a skill to the Hot tier only if it improves the set, otherwise it stays Warm.
(4)~At runtime, only Hot skills are injected into the agent's system prompt.
Rejected skills never cross the pre-commit boundary, so they cannot re-enter the next round's distillation context---i.e., $G$ is the last filter before defects propagate, and setting $G=\text{identity}$ recovers ungated evolution.
}
\label{fig:framework}
\end{figure*}

\section{Methodology}

We first formalize skill contamination in self-evolving agents as a taxonomy spanning three levels: individual, combinatorial, and systemic. We then show that once contamination is admitted, it becomes structurally irreversible: the lineage of derived skills makes post-hoc rollback unable to recover the full performance loss. Grounded in this irreversibility, we construct Verifier-as-Gatekeeper (VaG), a pre-commit gating mechanism that intercepts contamination before skills enter the agent's runtime context, organized as a progressive trust hierarchy with increasing verification stringency. Figure~\ref{fig:framework} illustrates the overall framework.

\paragraph{Problem Formalization and Contamination Taxonomy.}
Consider a self-evolving agent that distills reusable skills from its execution trajectories to accumulate capability.
Let $\mathcal{S}$ be the skill space; each skill $s \in \mathcal{S}$ is a structured experience unit written in natural language, carrying a short frontmatter (name, trigger condition, and body) followed by free-form guidance.
At evolution round $r$, the agent solves tasks using all previously accumulated skills $M_{r-1}$ as its decision context, then distills a set of new skills $S_r$ from that round's trajectories.
Existing self-evolution methods~\cite{xia2026metaclaw, wang2023voyager, lin2026agentic} share an implicit lifecycle assumption---every distilled skill is written to persistent storage and takes effect immediately---so the pool grows as
\begin{equation}
M_r = M_{r-1} \cup G(S_r), \quad S_r \sim \pi(\cdot \mid M_{r-1}),
\label{eq:evolution}
\end{equation}
where $\pi$ is the distillation process conditioned on $M_{r-1}$ and $G$ is an admission operator applied before new skills enter the pool.
Ungated evolution is the special case $G = \mathrm{identity}$, under which all of $S_r$ enters the decision context unconditionally.
Equation~\eqref{eq:evolution} exposes the vulnerability of this default: since $S_r$ is generated conditioned on $M_{r-1}$, any defective skill in $M_{r-1}$ can pass its flawed logic into $S_r$, propagating contamination across rounds.

To make this precise, let $\mathcal{D}_{\text{holdout}}$ be a held-out task set disjoint from the tasks used for distillation, and let $R(A, \tau)$ denote the agent's success on task $\tau$ when injected with skill set $A \subseteq \mathcal{S}$.
We write $R(A) = \mathbb{E}_{\tau \sim \mathcal{D}_{\text{holdout}}}[R(A, \tau)]$ for the aggregate success rate over $\mathcal{D}_{\text{holdout}}$, and define the \emph{injection gain} of $A$ over the current pool $M$ as
\begin{equation}
g(A, \tau) = R(M \cup A, \tau) - R(M, \tau).
\label{eq:injection_gain}
\end{equation}
Contamination then admits a three-level taxonomy.

\noindent\textbf{Individual contamination.}
A single skill $s$ is \emph{individually contaminating} on $\tau$ when its singleton gain is negative, $g(\{s\}, \tau) < 0$: injecting $s$ alone lowers the success rate.

\noindent\textbf{Combinatorial contamination.}
Individual harmlessness does not compose. There can exist skills $s_a, s_b$ with $g(\{s_a\}, \tau) \ge 0$ and $g(\{s_b\}, \tau) \ge 0$, yet
\begin{equation}
g(\{s_a, s_b\}, \tau) \;<\; 0 .
\label{eq:combinatorial}
\end{equation}
Detecting this requires \emph{joint} evaluation of candidate sets; independent per-skill judgment cannot see the interaction.

\noindent\textbf{Systemic contamination.}
The two effects surface macroscopically as a non-monotone capability curve.
Writing $P(k) = R(M_r)$ for the aggregate success rate when the pool has grown to $k = |M_r|$ skills, unconstrained evolution without admission control exhibits a critical size
\begin{equation}
k^* = \arg\max_{k} \, P(k),
\label{eq:critical}
\end{equation}
with $P(k)$ increasing for $k < k^*$ and decreasing for $k > k^*$.
This $k^*$ is the \emph{capability--contamination phase transition}: the point where the marginal value of adding skills turns from positive to negative.

\paragraph{Irreversibility of Contamination.}
A natural response to the tipping point is post-hoc remediation: identify the contaminating source skills and remove them from $M_r$.
Equation~\eqref{eq:evolution}, however, exposes a structural obstacle.
Because $S_r \sim \pi(\cdot \mid M_{r-1})$, skills distilled in later rounds are written with earlier skills as reference context.
Let $\text{desc}(s)$ be the set of skills derived from $s$ in subsequent rounds---those whose distillation context contained $s$.
Removing only the source $s$ leaves its descendants behind, so
\begin{equation}
R(M_r \setminus \{s\}) \;<\; R\bigl(M_r \setminus (\{s\} \cup \text{desc}(s))\bigr) .
\label{eq:irreversible}
\end{equation}
The left side is the recovery from removing the identified source alone; the right side is the best achievable by removing $s$ together with its entire lineage.
Inequality~\eqref{eq:irreversible} holds whenever $\text{desc}(s)$ contains at least one skill that inherited $s$'s flawed reasoning; in that regime source-only rollback is strictly dominated by full-lineage cleanup, and erasing the origin does not erase its consequences.

The design implication is direct: contamination \textbf{must be intercepted before skills enter the agent's runtime context}.
Pre-commit gating is not one option among several---it follows as a structural necessity from the irreversibility above.
We now construct such a mechanism.

\paragraph{Pre-commit Gating with a Progressive Trust Hierarchy.} Guided by the irreversibility result, we organize skill admission as a progressive trust hierarchy $\mathcal{L} = \{\text{Cold}, \text{Warm}, \text{Hot}\}$ with the partial order $\text{Cold} \prec \text{Warm} \prec \text{Hot}$.
Each newly distilled skill enters the Cold tier by default and does not
participate in the agent's decision context. Promotion between tiers is governed by gate predicates: skill $s$ is promoted from tier $\ell$ to $\ell'$ if and only if
\begin{equation}
G_{\ell \to \ell'}(s) = \top.
\label{eq:gate}
\end{equation}
The stringency of gate predicates increases with each tier, implementing the principle of progressively intensifying verification.

\textbf{Gate~1 (Cold $\to$ Warm): individual harmlessness.}
This gate asks whether a skill degrades performance when injected alone, and applies three complementary checks that a skill must all pass.
\begin{enumerate}
\item[(i)] \textbf{Structural validity.}
SchemaCritic verifies that $s$ satisfies a predefined frontmatter schema (all required fields are present and correctly typed).
This is a deterministic judgment independent of model reasoning:
$\nu_{\text{struct}}(s) \in \{0, 1\}$.

\item[(ii)] \textbf{Behavioral harmlessness.}
ExecCritic runs a single-skill A-B replay on $\mathcal{D}_{\text{holdout}}$, comparing an agent with $M \cup \{s\}$ against one with $M$ alone.
The skill passes when it does not degrade the aggregate rate, $R(M \cup \{s\}) \ge R(M)$, giving $\nu_{\text{exec}}(s) \in \{0, 1\}$.
We admit a skill that merely matches $M$ (rather than requiring strict improvement) because the estimate is noisy: a skill that is neutral here may still contribute in combination, and Gate~2 later prunes any that fail to help the set.

\item[(iii)] \textbf{Semantic consistency.}
AgentCritic evaluates via a single LLM call whether $s$ contains fabricated facts, logically contradicts existing skills in $M$, or recommends unsafe operations, outputting $\nu_{\text{sem}}(s) \in \{0, 1\}$.
\end{enumerate}
The Cold $\to$ Warm gate enforces conjunction:
$G_{\text{C} \to \text{W}}(s) = \nu_{\text{struct}} \land \nu_{\text{exec}}
\land \nu_{\text{sem}}$.
The three critics inspect different failure surfaces---format, observed behavior, and semantic content---and a skill is admitted only if it clears all three.
Because a harmful skill must evade every critic simultaneously to slip through, the conjunction is empirically far stricter than any single dimension: our ablations (Table~\ref{tab:ablation}) show each critic intercepts a largely disjoint class of harmful skills, so none substitutes for another.
This gate is also computationally economical: structural validation and
behavioral replay require no LLM calls, only semantic review consumes one
inference, and the first two filters eliminate the vast majority of candidates.

\textbf{Gate~2 (Warm $\to$ Hot): combinatorial safety via marginal-gain selection.}
The Cold $\to$ Warm gate certifies each skill in isolation, but not in company: two skills that each pass may still conflict once injected together.
The Warm $\to$ Hot transition is therefore a subset-selection problem---choose $H \subseteq W$ from the Warm pool $W$ to maximize joint held-out success.
Define the set utility function
\begin{equation}
f(H) = \mathbb{E}_{\tau \sim \mathcal{D}_{\text{holdout}}}\bigl[
       \,R(\text{agent} \oplus H, \tau)\,\bigr],
\label{eq:utility}
\end{equation}
where $\text{agent} \oplus H$ denotes injecting skill set $H$ into the agent's system prompt.
A direct solution requires enumerating $2^{|W|}$ subsets with separate replay, which is computationally prohibitive.
However, $f$ exhibits bounded submodularity over skill sets: when $H$ already contains several skills targeting the same functional domain (e.g., git operations), the marginal gain of adding another skill from that domain diminishes---prior skills already cover most scenarios in that domain.

Because skill injection can be contaminating (Eq.~\eqref{eq:injection_gain} may be negative), $f$ is not globally monotone: adding a skill can \emph{lower} it.
We therefore use \emph{marginal-gain greedy selection}: starting from the empty set, we repeatedly add the Warm candidate with the largest estimated joint gain and keep it only if it strictly improves measured held-out performance, stopping when no remaining candidate helps.
This is a heuristic rather than an algorithm with a worst-case optimality guarantee---the utility surface over skill sets is neither known to be submodular nor monotone---but it directly targets the combinatorial effect that per-skill checks cannot see, and it is cheap at our scale.
Each joint utility $f(H)$ is estimated as the mean of $k=3$ held-out replays.
Because Cold $\to$ Warm filtering leaves $|W| \le 15$ candidates per round, a single greedy pass costs at most $|W|-1$ joint replays, so we run it exactly without further approximation.

\section{Experiments}
\paragraph{Benchmarks.}

Our primary evaluation is on Terminal-Bench~2 (TB2)~\cite{merrill2026terminal}, a suite of hard, verifier-checked terminal tasks in which an agent operates a real shell inside a sandboxed container and its solution is graded by a deterministic checker.
Stratifying by TB2's official difficulty under a fixed seed, we split the tasks into three disjoint subsets: \emph{Event} (50 tasks) drives skill distillation and rolling evolution; \emph{Holdout} (14 tasks) serves only the A-B replays and marginal-gain estimates inside the gates; and \emph{Test} (25 tasks) is reserved for final evaluation, never touched by distillation or gating.
Because Event is the distillation split, its numbers are an optimistic upper bound, whereas Test is the honest held-out estimate.
To probe cross-benchmark transfer, we additionally evaluate on the full set of $200$ InterCode NL2Bash~\cite{yang2023intercode} tasks, which map natural-language instructions to single bash commands and share the shell ecosystem with TB2 while differing in task structure.

\paragraph{Agent and backbones.}
Each trial runs inside the Harbor TB2 runtime, which handles task dispatch, Docker orchestration, and deterministic scoring.
On top of it, the agent executes a hand-rolled ReAct loop: at each turn the language model emits either a shell command, which is run inside the per-trial container, or a termination action, and the observed output is appended to the context for the next turn.
The primary backbone driving distillation and gating is Hy3.
To test whether the evolved skills are tied to this backbone, the cross-model study freezes the skill pool and runs inference only on four further models spanning open- and closed-source families---DeepSeek-V4-Pro~\cite{xu2026deepseek}, GPT-5.4, Claude Sonnet~4.5, and Qwen3.6-35B-A3B~\cite{qwen36_35b_a3b}---with no re-evolution or gating on them.

\paragraph{Configurations.}
We compare four configurations under one agent implementation~\cite{xu2026agentica}.
\emph{Seed} injects three hand-written, general-purpose skills (\texttt{bash-essentials}, \texttt{file-ops}, \texttt{grep-find}) with no evolution, a static lower bound.
\emph{Ungated} admits every distilled skill unconditionally, matching the default assumption of Voyager and AHE.
\emph{Post-hoc Rollback}, applied after Ungated collapses at R5, removes only the harmful \emph{source} skills while keeping their descendants, testing whether cleanup can undo the irreversibility of Eq.~\eqref{eq:irreversible}.
\emph{VaG (ours)} runs the full Cold\,$\to$\,Warm\,$\to$\,Hot pipeline.

\paragraph{Implementation details.}
Evolution runs five rounds (R1--R5) on Event, with $k{=}3$ rollouts per task.
We report pass@1, and pass/fail is decided by TB2's built-in deterministic verifier without any LLM grader.
Remaining details are in the Supplementary Material.

\paragraph{Main Results.}
\begin{table*}[t]
\centering
\small
\setlength{\tabcolsep}{10pt}
\renewcommand{\arraystretch}{1.15}
\begin{tabular}{@{\extracolsep{\fill}}lccccccc@{}}
\toprule
Round & Pass@1 & $\Delta$Seed\,(pp) & Easy & Med & Hard & Pool & Tok/Trial\,(M) \\
\midrule
\multicolumn{8}{c}{\emph{\textbf{Seed} --- static baseline, no evolution}} \\
--- & 46\% & --- & 50\% & 52\% & 35\% & 3 & 0.35 \\
\midrule
\multicolumn{8}{c}{\emph{\textbf{Ungated} --- no admission control}} \\
R1 & 48\% & $+2$ & 100\% & 48\% & 41\% & 35 & 1.07 \\
R2 & 60\% & $+14$ & 100\% & 65\% & 53\% & 68 & 1.20 \\
R3 & \textbf{62\%} & $+16$ & 100\% & 65\% & 47\% & 105 & 1.30 \\
R4 & 52\% & $+6$ & 100\% & 52\% & 47\% & 141 & 1.24 \\
R5 & 50\% & $+4$ & 100\% & 55\% & 35\% & 179 & 1.19 \\
\;+\,Post-hoc Rollback & 52\% & $+6$ & 100\% & 58\% & 35\% & 171 & 1.15 \\
\midrule
\multicolumn{8}{c}{\emph{\textbf{VaG} --- pre-commit gated evolution (ours)}} \\
R1 & 52\% & $+6$ & 100\% & 55\% & 41\% & 5 & 0.77 \\
R2 & 58\% & $+12$ & 100\% & 61\% & 47\% & 15 & 0.82 \\
R3 & 62\% & $+16$ & 100\% & 68\% & 47\% & 25 & 0.87 \\
R4 & 68\% & $+22$ & 100\% & 74\% & 53\% & 30 & 0.90 \\
R5 & \textbf{72\%} & $+26$ & 100\% & 77\% & 59\% & 37 & 0.94 \\
\bottomrule
\end{tabular}
\caption{\textbf{Main results on Terminal-Bench~2 (Event split, 50 tasks).}
Pass@1 over five evolution rounds by TB2 difficulty tier (Easy: 2, Medium: 31, Hard: 17).
Pool is the number of skills in context; $\Delta$Seed is the gain over Seed, in points; Tok/Trial is mean tokens per trial. \textbf{Bold}: best pass@1 per block.}
\label{tab:main}
\end{table*}

\begin{figure}[t]
\centering
\includegraphics[width=\columnwidth]{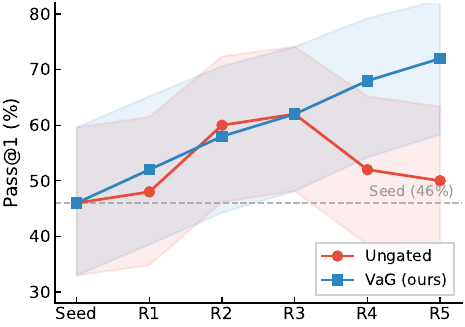}
\caption{\textbf{Pass@1 across five evolution rounds} (Event-50, $k{=}3$).
Red: Ungated; blue: VaG; dashed line: Seed baseline. Shaded bands are 95\% Wilson confidence intervals.}
\label{fig:phase_transition}
\end{figure}

We report pass@1 over five evolution rounds on Event-50 (Table~\ref{tab:main}, Figure~\ref{fig:phase_transition}), then verify the learned skills do not overfit this split through cross-model and cross-bench transfer on Test-25 (Tables~\ref{tab:cross_model}--\ref{tab:cross_bench}).

\emph{The phase transition is real and sharp.}
Seed holds constant at 46\%.
Ungated rises while its pool is small---48\% (R1, 35 skills) to 60\% (R2, 68 skills)---peaks at \textbf{62\%} (R3, 105 skills), then loses most of that gain: 52\% (R4, 141 skills), 50\% (R5, 179 skills), ending at R5 only 2pp above its own R1 despite growing from 35 to 179 skills.
This R3 peak instantiates the critical skill count $k^*$ of Eq.~\eqref{eq:critical}: beyond it, cumulative contamination outweighs the benefit of new skills.

\emph{Gating converts the inverted-U into monotone improvement.}
VaG rises every round---52\% (R1) $\to$ 72\% (R5)---always above Seed, with a Hot pool of 37 skills, one-fifth of Ungated's.
At R5, VaG exceeds Ungated by \textbf{22pp} and its \emph{best} round (R3, 62\%) by \textbf{10pp}: gating beats both unchecked accumulation and oracle early-stopping.

\emph{Contamination targets exactly the hard tasks.}
Ungated's Hard-tier pass@1 peaks early (53\% at R2) and erodes to 35\% at R5, while Medium falls from 65\% to 55\%.
VaG instead lifts Hard to 59\% at R5---a 24pp margin over Ungated---showing admission control protects precisely the multi-step tasks where contamination chains are longest.

\emph{Trajectory shape, not single-round significance, is the evidence.}
The 95\% Wilson CI bands in Figure~\ref{fig:phase_transition} (width $\approx$30pp, $n{=}50$, $k{=}3$) overlap at intermediate rounds, so per-round point tests are underpowered.
The claim rests on the \emph{qualitative divergence} of trajectories: Ungated is single-peaked about R3, VaG is monotone non-decreasing across all five rounds---a direction-of-trend difference 50 tasks cannot produce by chance.

\emph{Scale context.}
Recent self-evolution systems reach 68.9 (ACE~\cite{zhang2025agentic}) to 77.0 (AHE~\cite{lin2026agentic}) on the full TB2-89 suite with a stronger backbone and ten rounds.
Our goal differs: we isolate the contamination phenomenon, which calls for a small, controlled setup.
These are not head-to-head baselines: backbone, task suite, and round budget all differ.

\begin{figure}[t]
\centering
\includegraphics[width=0.9\columnwidth]{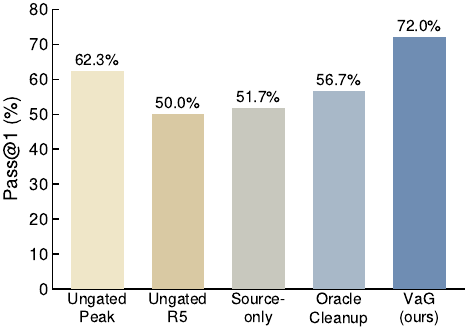}
\caption{\textbf{Residual contamination after post-hoc rollback} (Event-50, $k{=}3$).
Bars: pass@1 for the Ungated peak (R3), the collapsed R5 pool, source-only rollback, oracle full-lineage cleanup, and VaG~(R5).
Of the 12.3pp peak-to-R5 drop, source removal recovers only 1.7pp and oracle cleanup 6.7pp.}
\label{fig:rollback}
\end{figure}

\emph{Post-hoc cleanup cannot undo the damage.}
Removing the 8 harmful source skills from the collapsed R5 pool recovers only 2pp (50\%$\to$52\%); the residual 10pp gap to the R3 peak is locked in by descendants that inherited the contamination logic---the empirical signature of the irreversibility in Eq.~\eqref{eq:irreversible} (detailed below, Figure~\ref{fig:rollback}).
Admission control instead yields a compact pool whose capability grows every round, whereas unconditional accumulation buys a brief early gain and pays it back with interest.

\emph{Gating is cheap relative to what it prevents.}
Gate~1's first two checks use no LLM calls, so only the semantic critic costs one inference per candidate; Gate~2 reuses the Gate-1 replays, adding at most $|W|{-}1 \le 14$ joint replays per round.
This cost is paid once at admission, whereas a contaminant is paid for on \emph{every} later trial and, through its descendants, in every later round.
Accordingly, VaG's per-trial token cost stays low (0.77--0.94M) while Ungated's grows with its bloated pool (up to 1.30M) for worse accuracy---the gate buys accuracy and efficiency together.

\paragraph{Ablation Studies.}
We disable one VaG component at a time, keeping the rest fixed (Table~\ref{tab:ablation}).

\begin{table}[t]
\centering
\small
\setlength{\tabcolsep}{8pt}
\renewcommand{\arraystretch}{1.15}
\begin{tabular}{@{\extracolsep{\fill}}lccc@{}}
\toprule
Configuration & Pass@1 & Pool & $\Delta$VaG (pp) \\
\midrule
VaG (full) & \textbf{72\%} & 37 & --- \\
$-$\,Schema validation & 70\% & 37 & $-2$ \\
$-$\,Holdout replay & 62\% & 45 & \textbf{$-10$} \\
$-$\,Semantic check & 68\% & 40 & $-4$ \\
$-$\,Marginal-gain gate & 64\% & 58 & \textbf{$-8$} \\
\bottomrule
\end{tabular}
\caption{Component ablation on Event-50 (R5). $-$\,Marginal-gain gate admits every Gate-1 survivor to Hot without joint selection. $\Delta$VaG is the change from full VaG, in points. \textbf{Bold}: two largest drops.}
\label{tab:ablation}
\end{table}

Schema validation costs little ($-2$pp): malformed skills reach Warm but the backbone mostly ignores or repairs them, so format errors rarely translate into behavioral harm.
Holdout replay matters most ($-10$pp)---it is the only check that empirically tests behavior rather than surface form, so without it a skill that reads well but silently lowers the success rate passes Gate~1 undetected and directly raises the pool's contamination rate.
The semantic check costs $-4$pp, admitting fabricated or self-contradictory advice that executes without error yet misleads the agent.
Crucially, the three checks fail on \emph{different} skills---schema on malformed entries, replay on silently harmful ones, semantics on plausible-but-fabricated advice---so no single check substitutes for another, which is why Gate~1 requires all three rather than the strongest one alone.
Removing the marginal-gain gate is the clearest evidence of combinatorial contamination (Eq.~\eqref{eq:combinatorial}): with every Gate-1 survivor admitted, the Hot pool balloons from 37 to 58 skills yet pass@1 drops $-8$pp, showing that skills which each clear the individual checks can still conflict once injected together---exactly the interaction that per-skill judgment cannot detect.

\paragraph{Post-hoc Rollback Analysis.}
To test how much damage post-hoc removal can undo, we compare removing only the harmful \emph{source} skills from the collapsed Ungated R5 pool against an oracle that also removes their entire lineage---which requires provenance a real system lacks.

Figure~\ref{fig:rollback} decomposes the 12.3pp drop from the Ungated peak (62.3\%) to R5 (50.0\%): source removal recovers only 1.7pp (individual contamination), full lineage cleanup a further 5.0pp (combinatorial contamination via descendants), and the remaining 5.6pp is \textbf{irrecoverable} even under Oracle cleanup---no post-hoc operation can restore the pre-degradation state.
That even the oracle, with perfect provenance, leaves nearly half the drop unrecovered confirms that the loss is baked into the pool's state rather than carried by a few removable entries.
VaG~(72.0\%) exceeds Oracle by \textbf{15.3pp}, far beyond any recoverable gap, showing that preventing admission is decisively better than any cleanup after the fact.
As a concrete case, a git-conflict skill distilled at R3 seeded two derived skills at R4 (merge and rebase workflows); after source-only rollback both remained and kept failing 4 of 7 git-related Test-25 tasks (full chain in the Supplementary Material).
Gating must therefore act before skills enter the runtime context.

\paragraph{Cross-Model Generalization.}
To show the skills are not fitted to Event-50 or to Hy3, we freeze the R5 Hot pool and evaluate it on Test-25 with other backbones, without re-evolution.
The frozen pool gives positive lift on all five backbones ($+8$ to $+16$pp; Table~\ref{tab:cross_model}).
GPT-5.4 and Claude Sonnet~4.5 reach 56\%, above Hy3's own 44\%, indicating the gate-filtered skills encode model-agnostic engineering knowledge in natural language, from which stronger backbones extract more.

\begin{table}[t]
\centering
\small
\setlength{\tabcolsep}{8pt}
\renewcommand{\arraystretch}{1.15}
\begin{tabular}{@{\extracolsep{\fill}}lccc@{}}
\toprule
Backbone & Seed & VaG Hot & $\Delta$ (pp) \\
\midrule
Hy3 & 32\% & 44\% & $+12$ \\
DeepSeek-V4-Pro & 32\% & 40\% & $+8$ \\
GPT-5.4 & 44\% & \textbf{56\%} & $+12$ \\
Claude Sonnet~4.5 & 48\% & \textbf{56\%} & $+8$ \\
Qwen3.6-35B-A3B & 36\% & 52\% & \textbf{$+16$} \\
\bottomrule
\end{tabular}
\caption{Cross-model transfer on Test-25. The Hy3-evolved Hot pool (R5) is frozen and evaluated on other backbones without re-evolution.}
\label{tab:cross_model}
\end{table}

\paragraph{Cross-Benchmark Generalization.}
To confirm the skills capture general shell knowledge rather than TB2-specific patterns, we evaluate the frozen R5 pools on InterCode NL2Bash~\cite{yang2023intercode}, which maps natural-language instructions to bash commands.
It shares the shell ecosystem with TB2 but tests single-step mapping rather than multi-step operations, giving a domain-overlapping yet format-distinct test.

\begin{table}[t]
\centering
\small
\setlength{\tabcolsep}{8pt}
\renewcommand{\arraystretch}{1.15}
\begin{tabular}{@{\extracolsep{\fill}}lccc@{}}
\toprule
Configuration & Pool & Pass@1 & $\Delta$Seed\,(pp) \\
\midrule
Seed & 3 & 57.5\% & --- \\
Ungated (R5) & 179 & 65.5\% & $+8.0$ \\
VaG (R5) & 37 & \textbf{69.0\%} & \textbf{$+11.5$} \\
\bottomrule
\end{tabular}
\caption{Cross-benchmark transfer to InterCode NL2Bash. Frozen R5 pools, no re-evolution. \textbf{Bold}: best.}
\label{tab:cross_bench}
\end{table}

The comparison direction is preserved: VaG's 37 Hot skills reach 69.0\%, above Ungated's 179 skills at 65.5\% and Seed at 57.5\%.
Ungated stays net-positive over Seed here ($+8.0$pp): on these short tasks each trial invokes few skills and contamination chains stay shallow, so new skills still help---consistent with contamination biting hardest on long, multi-step tasks.
Gate-filtered skills thus carry reusable shell-engineering knowledge across benchmarks, not TB2-specific heuristics.

\section{Conclusion}
We identified a non-monotonic capability trajectory in self-evolving agents---accumulated skills first help, then contaminate the decision context---and showed this contamination is structurally irreversible, since post-hoc removal of a source skill cannot undo the flawed reasoning its descendants inherit.
This makes skill admission a pre-commit necessity, which we address with Verifier-as-Gatekeeper (VaG): three complementary critics filter each skill, and a marginal-gain selection removes combinatorial conflicts.
On Terminal-Bench~2, VaG improves every round to 72\% pass@1 with a $5\times$ smaller pool, while ungated evolution peaks and collapses; the frozen pool also transfers across five backbones and to a second benchmark.
Pre-commit verification is thus a structural requirement, not an optional refinement, for reliable self-evolving systems.

\bibliography{aaai2027}

@article{wang2023voyager,
  title={Voyager: An open-ended embodied agent with large language models},
  author={Wang, Guanzhi and Xie, Yuqi and Jiang, Yunfan and Mandlekar, Ajay and Xiao, Chaowei and Zhu, Yuke and Fan, Linxi and Anandkumar, Anima},
  journal={arXiv preprint arXiv:2305.16291},
  year={2023}
}

@article{lin2026agentic,
  title={Agentic harness engineering: Observability-driven automatic evolution of coding-agent harnesses},
  author={Lin, Jiahang and Liu, Shichun and Pan, Chengjun and Lin, Lizhi and Dou, Shihan and Xi, Zhiheng and Huang, Xuanjing and Yan, Hang and Han, Zhenhua and Gui, Tao and others},
  journal={arXiv preprint arXiv:2604.25850},
  year={2026}
}

@article{zhang2025agentic,
  title={Agentic context engineering: Evolving contexts for self-improving language models},
  author={Zhang, Qizheng and Hu, Changran and Upasani, Shubhangi and Ma, Boyuan and Hong, Fenglu and Kamanuru, Vamsidhar and Rainton, Jay and Wu, Chen and Ji, Mengmeng and Li, Hanchen and others},
  journal={arXiv preprint arXiv:2510.04618},
  year={2025}
}

@article{xia2026metaclaw,
  title={MetaClaw: Just Talk--An Agent That Meta-Learns and Evolves in the Wild},
  author={Xia, Peng and Chen, Jianwen and Yang, Xinyu and Tu, Haoqin and Liu, Jiaqi and Xiong, Kaiwen and Han, Siwei and Qiu, Shi and Ji, Haonian and Zhou, Yuyin and others},
  journal={arXiv preprint arXiv:2603.17187},
  year={2026}
}

@article{nemhauser1978analysis,
  title={An analysis of approximations for maximizing submodular set functions—I},
  author={Nemhauser, George L and Wolsey, Laurence A and Fisher, Marshall L},
  journal={Mathematical programming},
  volume={14},
  number={1},
  pages={265--294},
  year={1978},
  publisher={Springer}
}

@article{das2011submodular,
  title={Submodular meets spectral: Greedy algorithms for subset selection, sparse approximation and dictionary selection},
  author={Das, Abhimanyu and Kempe, David},
  journal={arXiv preprint arXiv:1102.3975},
  year={2011}
}

@article{mirzasoleiman2014lazier,
  title={Lazier than lazy greedy},
  author={Mirzasoleiman, Baharan and Badanidiyuru, Ashwinkumar and Karbasi, Amin and Vondr{\'a}k, Jan and Krause, Andreas},
  journal={arXiv preprint arXiv:1409.7938},
  year={2014}
}

@article{madaan2023self,
  title={Self-refine: Iterative refinement with self-feedback},
  author={Madaan, Aman and Tandon, Niket and Gupta, Prakhar and Hallinan, Skyler and Gao, Luyu and Wiegreffe, Sarah and Alon, Uri and Dziri, Nouha and Prabhumoye, Shrimai and Yang, Yiming and others},
  journal={Advances in neural information processing systems},
  volume={36},
  pages={46534--46594},
  year={2023}
}

@article{shinn2023reflexion,
  title={Reflexion: Language agents with verbal reinforcement learning},
  author={Shinn, Noah and Cassano, Federico and Gopinath, Ashwin and Narasimhan, Karthik and Yao, Shunyu},
  journal={Advances in neural information processing systems},
  volume={36},
  pages={8634--8652},
  year={2023}
}

@article{merrill2026terminal,
  title={Terminal-bench: Benchmarking agents on hard, realistic tasks in command line interfaces},
  author={Merrill, Mike A and Shaw, Alexander G and Carlini, Nicholas and Li, Boxuan and Raj, Harsh and Bercovich, Ivan and Shi, Lin and Shin, Jeong Yeon and Walshe, Thomas and Buchanan, E Kelly and others},
  journal={arXiv preprint arXiv:2601.11868},
  year={2026}
}

@inproceedings{zhao2024expel,
  title={Expel: Llm agents are experiential learners},
  author={Zhao, Andrew and Huang, Daniel and Xu, Quentin and Lin, Matthieu and Liu, Yong-Jin and Huang, Gao},
  booktitle={Proceedings of the AAAI Conference on Artificial Intelligence},
  volume={38},
  number={17},
  pages={19632--19642},
  year={2024}
}

@article{wang2024agent,
  title={Agent workflow memory},
  author={Wang, Zora Zhiruo and Mao, Jiayuan and Fried, Daniel and Neubig, Graham},
  journal={arXiv preprint arXiv:2409.07429},
  year={2024}
}

@article{yang2023intercode,
  title={Intercode: Standardizing and benchmarking interactive coding with execution feedback},
  author={Yang, John and Prabhakar, Akshara and Narasimhan, Karthik and Yao, Shunyu},
  journal={Advances in Neural Information Processing Systems},
  volume={36},
  pages={23826--23854},
  year={2023}
}

@article{tao2024survey,
  title={A survey on self-evolution of large language models},
  author={Tao, Zhengwei and Lin, Ting-En and Chen, Xiancai and Li, Hangyu and Wu, Yuchuan and Li, Yongbin and Jin, Zhi and Huang, Fei and Tao, Dacheng and Zhou, Jingren},
  journal={arXiv preprint arXiv:2404.14387},
  year={2024}
}

@inproceedings{yao2022react,
  title={React: Synergizing reasoning and acting in language models},
  author={Yao, Shunyu and Zhao, Jeffrey and Yu, Dian and Shafran, Izhak and Narasimhan, Karthik R and Cao, Yuan},
  booktitle={NeurIPS 2022 Foundation Models for Decision Making Workshop},
  year={2022}
}

@inproceedings{park2023generative,
  title={Generative agents: Interactive simulacra of human behavior},
  author={Park, Joon Sung and O'Brien, Joseph and Cai, Carrie Jun and Morris, Meredith Ringel and Liang, Percy and Bernstein, Michael S},
  booktitle={Proceedings of the 36th annual acm symposium on user interface software and technology},
  pages={1--22},
  year={2023}
}

@article{feige2011maximizing,
  title={Maximizing non-monotone submodular functions},
  author={Feige, Uriel and Mirrokni, Vahab S and Vondr{\'a}k, Jan},
  journal={SIAM Journal on Computing},
  volume={40},
  number={4},
  pages={1133--1153},
  year={2011},
  publisher={SIAM}
}

@article{packer2023memgpt,
  title={MemGPT: towards LLMs as operating systems.},
  author={Packer, Charles and Fang, Vivian and Patil, Shishir\_G and Lin, Kevin and Wooders, Sarah and Gonzalez, Joseph\_E},
  year={2023},
  publisher={ArXiv}
}

@article{zhang2022automatic,
  title={Automatic chain of thought prompting in large language models},
  author={Zhang, Zhuosheng and Zhang, Aston and Li, Mu and Smola, Alex},
  journal={arXiv preprint arXiv:2210.03493},
  year={2022}
}

@article{schick2023toolformer,
  title={Toolformer: Language models can teach themselves to use tools},
  author={Schick, Timo and Dwivedi-Yu, Jane and Dess{\`\i}, Roberto and Raileanu, Roberta and Lomeli, Maria and Hambro, Eric and Zettlemoyer, Luke and Cancedda, Nicola and Scialom, Thomas},
  journal={Advances in neural information processing systems},
  volume={36},
  pages={68539--68551},
  year={2023}
}

@article{wang2024survey,
  title={A survey on large language model based autonomous agents},
  author={Wang, Lei and Ma, Chen and Feng, Xueyang and Zhang, Zeyu and Yang, Hao and Zhang, Jingsen and Chen, Zhiyuan and Tang, Jiakai and Chen, Xu and Lin, Yankai and others},
  journal={Frontiers of Computer Science},
  volume={18},
  number={6},
  pages={186345},
  year={2024},
  publisher={Springer}
}

@inproceedings{huang2023large,
  title={Large language models can self-improve},
  author={Huang, Jiaxin and Gu, Shixiang and Hou, Le and Wu, Yuexin and Wang, Xuezhi and Yu, Hongkun and Han, Jiawei},
  booktitle={Proceedings of the 2023 conference on empirical methods in natural language processing},
  pages={1051--1068},
  year={2023}
}

@article{zelikman2022star,
  title={Star: Bootstrapping reasoning with reasoning},
  author={Zelikman, Eric and Wu, Yuhuai and Mu, Jesse and Goodman, Noah},
  journal={Advances in Neural Information Processing Systems},
  volume={35},
  pages={15476--15488},
  year={2022}
}

@article{gulcehre2023reinforced,
  title={Reinforced self-training (rest) for language modeling},
  author={Gulcehre, Caglar and Paine, Tom Le and Srinivasan, Srivatsan and Konyushkova, Ksenia and Weerts, Lotte and Sharma, Abhishek and Siddhant, Aditya and Ahern, Alex and Wang, Miaosen and Gu, Chenjie and others},
  journal={arXiv preprint arXiv:2308.08998},
  year={2023}
}

@misc{mei2026searcharttraininglonghorizonsearch,
      title={SearchArt: Training Long-Horizon Search Agent with Scalable Synthetic and Verified Task}, 
      author={Lang Mei and Xiaohan Yu and Chong Chen and Liyan Liu and Xiangnan Chen and Jinchao Ma and Chao Feng and Li Huang and Siyu Mo and Sichen Kang and Yunkun Xu and Zhihan Yang and Zhujun Xue and Jingren Zhang and Qing He and Yingdi Huang and Hao Jiang and Ziao Ma and Zewei Pan and Minhao Sun and Zhuo Tao and Jinzhao Xiao and Gangtao Xin and Huanyao Zhang and Wenjian Zhang and Jiangshan Zhang and Guojie Zhu and Jiaxin Mao and Wentao Zhang},
      year={2026},
      eprint={2607.24850},
      archivePrefix={arXiv},
      primaryClass={cs.IR},
      url={https://arxiv.org/abs/2607.24850}, 
}

@article{skillops,
  author       = {Hongji Pu and
                  Xinyuan Song and
                  Liang Zhao},
  title        = {SkillOps: Managing {LLM} Agent Skill Libraries as Self-Maintaining
                  Software Ecosystems},
  journal      = {CoRR},
  volume       = {abs/2605.13716},
  year         = {2026},
  url          = {https://doi.org/10.48550/arXiv.2605.13716},
  doi          = {10.48550/ARXIV.2605.13716},
  eprinttype   = {arXiv},
  eprint       = {2605.13716},
  bibsource    = {dblp computer science bibliography, https://dblp.org}
}

@article{routeguard,
  author       = {Wenjie Xiao and
                  Xuehai Tang and
                  Biyu Zhou and
                  Songlin Hu and
                  Jizhong Han},
  title        = {RouteGuard: Internal-Signal Detection of Skill Poisoning in {LLM}
                  Agents},
  journal      = {CoRR},
  volume       = {abs/2604.22888},
  year         = {2026},
  url          = {https://doi.org/10.48550/arXiv.2604.22888},
  doi          = {10.48550/ARXIV.2604.22888},
  eprinttype   = {arXiv},
  eprint       = {2604.22888},
  bibsource    = {dblp computer science bibliography, https://dblp.org}
}

@article{evolver,
  author       = {Rong Wu and
                  Xiaoman Wang and
                  Jianbiao Mei and
                  Pinlong Cai and
                  Daocheng Fu and
                  Cheng Yang and
                  Licheng Wen and
                  Xuemeng Yang and
                  Yufan Shen and
                  Yuxin Wang and
                  Botian Shi},
  title        = {EvolveR: Self-Evolving {LLM} Agents through an Experience-Driven Lifecycle},
  journal      = {CoRR},
  volume       = {abs/2510.16079},
  year         = {2025},
  url          = {https://doi.org/10.48550/arXiv.2510.16079},
  doi          = {10.48550/ARXIV.2510.16079},
  eprinttype   = {arXiv},
  eprint       = {2510.16079},
  bibsource    = {dblp computer science bibliography, https://dblp.org}
}

@inproceedings{sage,
  author       = {Jiongxiao Wang and
                  Qiaojing Yan and
                  Yawei Wang and
                  Yijun Tian and
                  Soumya Smruti Mishra and
                  Zhichao Xu and
                  Megha Gandhi and
                  Panpan Xu and
                  Lin Lee Cheong},
  editor       = {Maria Liakata and
                  Viviane P. Moreira and
                  Jiajun Zhang and
                  David Jurgens},
  title        = {Reinforcement Learning for Self-Improving Agent with Skill Library},
  booktitle    = {Proceedings of the 64th Annual Meeting of the Association for Computational
                  Linguistics (Volume 1: Long Papers), {ACL} 2026, San Diego, California,
                  United States, July 2-7, 2026},
  pages        = {1529--1550},
  publisher    = {Association for Computational Linguistics},
  year         = {2026},
  url          = {https://aclanthology.org/2026.acl-long.69/},
  bibsource    = {dblp computer science bibliography, https://dblp.org}
}

@article{xu2026deepseek,
  title={Deepseek-v4: Towards highly efficient million-token context intelligence},
  author={Xu, Anyi and Lin, Bangcai and Xue, Bing and Wang, Bingxuan and Xu, Bingzheng and Wu, Bochao and Zhang, Bowei and Lin, Chaofan and Dong, Chen and Ling, Chenchen and others},
  journal={arXiv preprint arXiv:2606.19348},
  year={2026}
}

@misc{qwen36_35b_a3b,
    title = {{Qwen3.6-35B-A3B}: Agentic Coding Power, Now Open to All},
    url = {https://qwen.ai/blog?id=qwen3.6-35b-a3b},
    author = {{Qwen Team}},
    month = {April},
    year = {2026}
}

@misc{xu2026agentica,
  title  = {{Agentica}: A Human-Centric Framework for Large Language Model Agent Workflows},
  author = {Xu, Ming},
  url    = {https://github.com/shibing624/agentica},
  year   = {2026}
}

\end{document}